\pdfoutput=1

\documentclass[11pt]{article}
\usepackage[]{acl}

\usepackage{times}
\usepackage{latexsym}

\usepackage[T1]{fontenc}

\usepackage[utf8]{inputenc}

\usepackage{microtype}

\usepackage{subfigure}
\usepackage{amssymb}
\usepackage{stfloats}
\usepackage{algorithm}
\usepackage{algorithmic}
\usepackage{float}
\usepackage{graphicx}
\usepackage{amsmath}
\usepackage{booktabs}
\usepackage{hyperref}
\usepackage{inconsolata}

\usepackage{enumitem}
\usepackage{multirow}
\newtheorem{myDef}{Definition}
\newtheorem{myThrm}{Theorem}
\newenvironment{myproof}{{\noindent\it Proof.}\noindent}{\hfill $\square$\par}
\title{HILL: Hierarchy-aware Information Lossless Contrastive Learning for Hierarchical Text Classification}
\author{
	He Zhu$^{1*}$, Junran Wu$^{1*}$,  Ruomei Liu$^1$,  Yue Hou$^1$, Ze Yuan$^1$, Shangzhe Li$^3$,\\
	 \textbf{Yicheng Pan$^{1,2\dag}$, and Ke Xu$^{1,2}$}  \\
	 	$^1$State Key Lab of Software Development Environment,\\
	 	Beihang University, Beijing, 100191, China\\
	 	$^{2}$Zhongguancun Laboratory, Beijing, 100094, China \\
	 	$^{3}$School of Statistics and Mathematics,\\
	 	Central University of Finance and Economics, Beijing 100081, China \\
	 	\{roy\_zh, wu\_junran, rmliu, hou\_yue, yuanze1024, yichengp, kexu\}@buaa.edu.cn, \\shangzheli@cufe.edu.cn}
\begin{document}
\maketitle
\begin{abstract}
Existing self-supervised methods in natural language processing (NLP), especially hierarchical text classification (HTC), mainly focus on self-supervised contrastive learning, extremely relying on human-designed augmentation rules to generate contrastive samples, which can potentially corrupt or distort the original information. In this paper, we tend to investigate the feasibility of a contrastive learning scheme in which the semantic and syntactic information inherent in the input sample is adequately reserved in the contrastive samples and fused during the learning process. Specifically, we propose an information lossless contrastive learning strategy for HTC, namely \textbf{H}ierarchy-aware \textbf{I}nformation \textbf{L}ossless contrastive \textbf{L}earning (HILL), which consists of a text encoder representing the input document, and a structure encoder directly generating the positive sample. The structure encoder takes the document embedding as input, extracts the essential syntactic information inherent in the label hierarchy with the principle of structural entropy minimization, and injects the syntactic information into the text representation via hierarchical representation learning. Experiments on three common datasets are conducted to verify the superiority of HILL.
\end{abstract}
\renewcommand{\thefootnote}{}
\footnotetext{$^{*}$Equal Contribution.}
\footnotetext{$^{\dag}$Correspondence to: Yicheng Pan.}
\section{Introduction}
Self-supervised learning (SSL) has exhibited remarkable success across various domains in deep learning, empowering models with potent representation capabilities. Based on these achievements, researchers have incorporated contrastive learning into hierarchical text classification \citep{HGCLR}, a challenging sub-task within the realm of text multi-label classification. Beyond processing text samples, hierarchical text classification methods should handle a predefined directed acyclic graph in the corpus, referred to as the label hierarchy. While language models such as BERT \citep{BERT} are pretrained on textual data, their efficacy in handling structural input is limited. Consequently, researchers have introduced a Graph Neural Network (GNN)-based encoder to establish a dual-encoder framework for HTC \citep{HiAGM, HTCInfoMax, HiMatch}.

\begin{figure}[!t]
	\subfigure[Previous Methods\label{fig:prev}]{\includegraphics[width=0.23\textwidth]{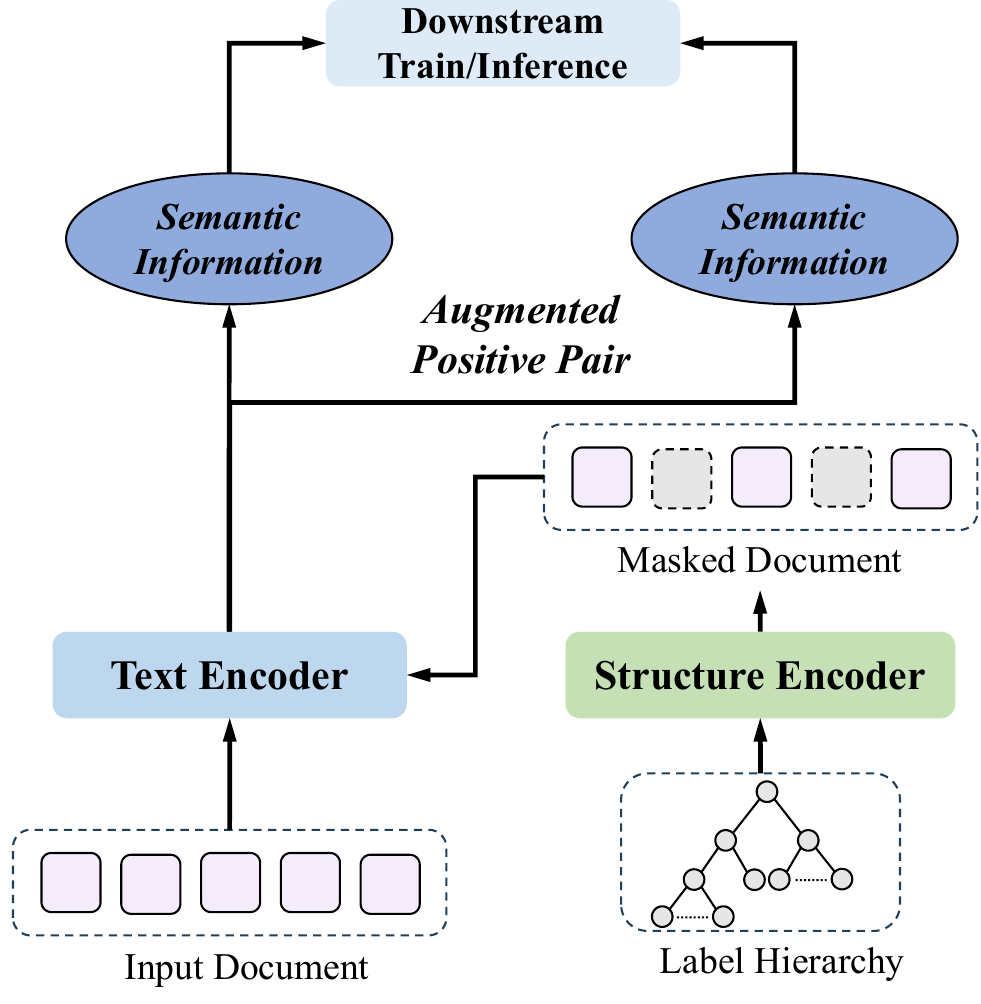}}
	\subfigure[Ours\label{fig:ours}]{\includegraphics[width=0.23\textwidth]{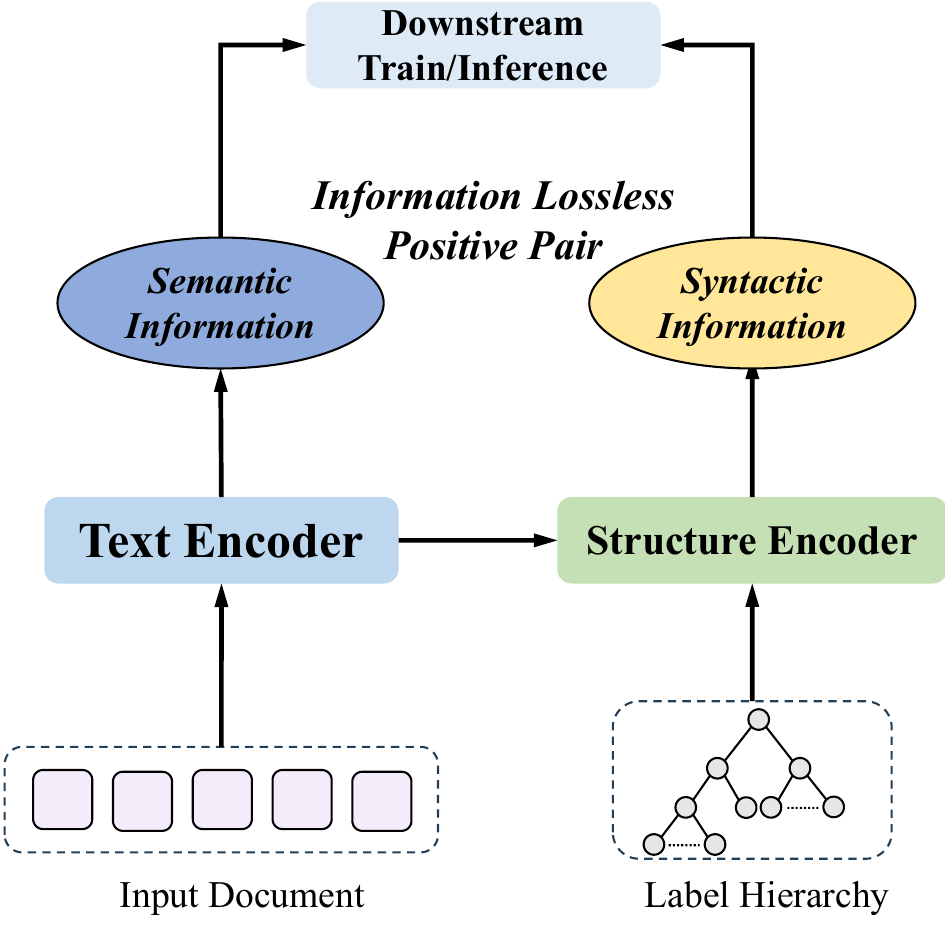}}\\
	\caption{Comparison between HILL and previous methods. (a) Previous work use structure encoder in data augmentation. (b) Our method extracting syntactic information in information lossless learning paradigm.}
\end{figure}
Although the structure encoder contributes to representing the label hierarchy, the dual-encoder framework simply blends the outputs of encoders. In an effort to integrate the label hierarchy into BERT, \citet{HGCLR} propose a contrastive learning framework, in which BERT functions as a siamese network \citep{siamese} accepting both the raw text and the masked text, where the mask is generated by the structure encoder. However, their contrastive learning process essentially relies on data augmentation, even with the involvement of the structure encoder in the masking process, as depicted in Figure~\ref{fig:prev}. According to the data processing inequality \citep{Cover2006}, applying data augmentation to the raw text may potentially erase sufficient semantic information relevant to downstream prediction.

To maximally preserve the semantic information in the text and effectively leverage the structural encoder in the contrastive learning process for HTC, we tend to inject the essential information inherent in the label hierarchy into text representations rather than augmenting the input document. As shown in Figure~\ref{fig:ours}, our structure encoder directly generates feature vectors by fusing textual and structural information, in contrast to masking the text as illustrated in Figure~\ref{fig:prev}. Following the insights of \citet{Li2016StructuralEntropy}, where structural entropy encodes and decodes the essential structure of the original system to support its semantic analysis. Since hierarchical text classification holds both semantic and syntactic information, we aim to design our model with the guidance of structural entropy. Specifically, we implement a suite of algorithms to minimize the structural entropy of label hierarchies by constructing their coding trees. Subsequently, we design a structure encoder to perform representation learning on coding trees, in which the leaf-node embeddings are initialized by the text encoder while the non-leaf node embeddings are iteratively obtained from bottom to top. Afterward, the structure encoder generates an overall representation of the coding tree, which serves as a contrastive sample for the text encoder. Additionally, we provide a definition of information lossless learning and prove that the information retained by our approach is the upper bound of any augmented data. In comparison with other contrastive and supervised learning methods, our model achieves significant performance gains on three common datasets. Overall, the contributions of our work can be summarized as follows:
\begin{itemize}
\item To realize information lossless learning, we decode the essential structure of label hierarchies through the proposed algorithms under the guidance of structural entropy, supporting the semantic analysis for HTC.
\vspace{-2mm}
\item We propose a contrastive learning framework, namely HILL, which fuses the structural information from the label hierarchies into the given document embeddings, while the semantic information from the input document is maximally preserved.
\vspace{-2mm}
\item We define information lossless learning for HTC and prove that the information retained by HILL is the upper bound of any other augmentation-based methods.
\vspace{-2mm}
\item Experiments conducted on three common datasets demonstrate the effectiveness and efficiency of HILL. For reproducibility, source code is available at \href{https://github.com/Rooooyy/HiTIN}{https://github.com/Rooooyy/HILL}.
\end{itemize}
\section{Related Works}
\paragraph{Hierarchical Text Classification.}
Existing works for HTC could be categorized into local and global approaches \citep{HiAGM}. Local approaches build multiple models for labels in different levels in the hierarchy, conveying the information from models in the upper levels to those in the bottom \citep{HDLTex, Shimura2018HFTCNNLH, Banerjee2019HierarchicalTL, Huang2019HierarchicalMT}. On the contrary, global studies treat HTC as a flat multi-label classification problem\citep{Gopal2013RecursiveRF, You2019AttentionXML, Aly2019HierarchicalMC, Mao2019HiRL, Wu2019Meta, Rojas2020}.  

Recently, \citet{HiAGM} introduce a dual-encoder framework consisting of a text and a graph encoder to separately handle the text and the label hierarchy. Based on HiAGM \citep{HiAGM}, \citet{Chen2020HyperbolicIM} jointly model the text and labels in the hyperbolic space. \citet{HiMatch} formulate HTC as a semantic matching problem. \citet{HTCInfoMax} introduce information maximization to capture the interaction between text and label while erasing irrelevant information. 

Given the success of Pretrained Language Models (PLMs), researchers attempt to utilize their powerful abilities in HTC. \citet{HGCLR} propose a contrastive learning framework for HTC to make BERT learn from the structure-encoder-controlled text augmentation. \citet{HPT} introduce prompt tuning and construct dynamic templates for HTC. \citet{HBGL} encode the global hierarchies with BERT, extract the local hierarchies, and feed them into BERT in a prompt-tuning-like schema. Despite their success, neither existing contrastive learning nor prompt tuning methods try to improve the structure encoder.

\paragraph{Contrastive Learning.}
Inspired by the pretext tasks in GPT \citep{GPT1} and BERT \citep{BERT}, researchers originally proposed contrastive learning in computer vision \citep{2020simclr, MOCO}, addressing the limitations of previous methods in training with massive unlabeled visual data. Numerous studies have shown that the key to contrastive learning lies in the construction of positive pairs \citep{Tian2020, Caron2020, Jaiswal2020, Grill2020}, especially in natural language processing \citep{Wu2020,  Yan2021, Meng2021,  Pan2022}. 
\paragraph{Structural Entropy.}
Structural entropy \citep{Li2016StructuralEntropy} is a natural extension of Shannon entropy \citep{Shannon} on a structural system, which could measure the structure complexity of the system. Non-Euclidean data, especially graph data, is a typical structured system. The structural entropy of a graph is defined as the average length of the codewords obtained by a random walk under a specific coding scheme, namely coding tree \citep{Li2016StructuralEntropy}. In the past few years, structural entropy has been successfully applied in community security \citep{Liu2019} graph classification \citep{Wu2022HRN, Yang2023}, text classification \citep{Zhang2022HINT}, graph pooling \citep{wu2022structural}, graph contrastive learning \citep{wu2023sega}, and node clustering \citep{Wang2023a}.
\begin{figure*}[!th]
	\centering
	\includegraphics[width=\textwidth]{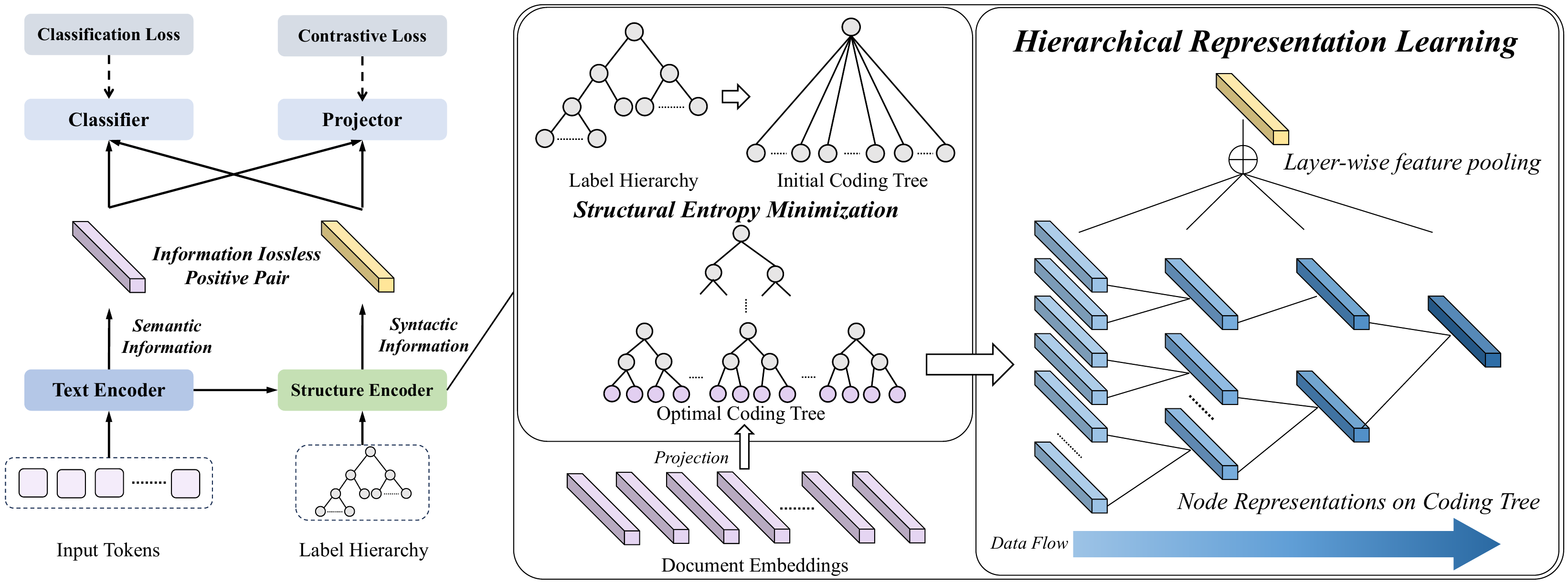}
	\caption{\label{fig:model}An example of our model with $K=3$. We first feed the document $D$ into the text encoder to extract the semantic information. Next, the structure encoder takes label hierarchy $G_L$ as input and constructs the optimal coding tree $T_L$ with Algorithm~\ref{alg:1} under the guidance of structural entropy. In the hierarchical representation learning module, the leaf node embeddings are initialized by the document embeddings, and the representations of non-leaf nodes are learned from bottom to top. The structure encoder finally generates an information lossless positive view for the text encoder, which is formulated in Section~\ref{sec:proof} and proved in Appendix~\ref{apdx:proof}.}
	\vspace{-2mm}
\end{figure*}
\section{Methodology}
In this section, we first give a problem definition of hierarchical text classification. Next, we elaborate on the working process of the text encoder (in Section~\ref{sec:text_encoder}) and the structure encoder (in Section~\ref{sec:structure_encoder}) of the proposed HILL. Theoretical analysis is further given to reveal the information lossless property of HILL for HTC in Section~\ref{sec:proof}. Overall, the framework of HILL is shown in Figure~\ref{fig:model}.
\subsection{Problem Definition}
In hierarchical text classification, the label set is predefined and represented as a directed acyclic graph, namely the label hierarchy. Every label that appears in the corpus corresponds to a unique node on the hierarchy.  Each non-root node is pointed by only one node in the upper levels, i.e. its parent node. In the ground-truth label set $\mathcal{Y}$ of any sample, a non-root label $y_i$ always co-occurs with its parent nodes, put differently, for any $y_i \in \mathcal{Y}$, the parent node of $y_i$ is also in $\mathcal{Y}$.  Given a document $D$ to be classified, where $D=\{w_1, w_2, \dots, w_N\}$ is commonly treated as a sequence with $N$ tokens, an HTC model should predict a subset  $\hat{\mathcal{Y}}$ of the complete label set $\mathbb{Y}$. 
\subsection{Text Encoder}
\label{sec:text_encoder}
Our framework is compatible with multiple document representation models. To maintain consistency with previous works, we utilize BERT \citep{BERT} as the text encoder. 

First, the input document $D$ is tokenized into a sequence with $N$ tokens and then padded with two special tokens:
\begin{equation}
	\tilde{D} = \{[CLS], w_1, w_2, \dots, w_N, [SEP]\},
\end{equation}
where $[CLS]$ and $[SEP]$ are respectively recognized as the beginning and the end of the document. 

Next, the BERT encoder takes the padded document $\tilde{D}$ as input and generates  hidden representations of each token, formally, $H_{BERT} = \mathcal{F}_{BERT}(\tilde{D}),$
where $H_{BERT} \in \mathbb{R}^{(N+2) \times d_B} $ is the token embedding matrix while $\mathcal{F}_{BERT}(\cdot)$ denotes the holistic BERT model. Afterward, the $[CLS]$ embedding is taken as the representation of the entire document. That is, $h_D = H_{BERT}^{[CLS]} = H_{BERT}^0,$ where $h_D \in \mathbb{R}^{d_B}$ is the document embedding, and $d_B$ is the hidden size of BERT.
\subsection{Structure Encoder}
\label{sec:structure_encoder}
 To implement information lossless contrastive learning in the structure encoder, we propose an algorithm to extract structural information from the label hierarchy via structural entropy \citep{Li2016StructuralEntropy} minimization and a hierarchical representation learning mechanism to inject the structural information into text embeddings. Thereafter, the structure encoder generates a positive view of the document that retains both semantic and structural information losslessly.
\paragraph{Structural Entropy.}
 In \citet{Li2016StructuralEntropy}, the structural entropy of a graph $G=(V_G, E_G)$ is defined as the average length of the codewords obtained by a random walk under a specific coding pattern named coding tree, formally,
\begin{equation}
	H^{T}(G)=-\sum_{v \in T} \frac{g_{v}}{vol(G)}\log{\frac{vol(v)}{vol(v^+)}},
\end{equation}
where $v$ is a non-root node of coding tree $T$ which represents a subset of $V_G$, $v^+$ is the parent node of $v$ on the coding tree. $g_{v}$ represents the number of $v$'s cut edges on $G$. $vol(G)$ denotes the volume of graph $G$ while $vol(v)$ and $vol(v^+)$ is the sum of the degree of nodes partitioned by $v$ and $v^+$. 

The height of the coding tree should be fixed to formulate a certain coding scheme. Therefore, the $K$-dimensional structural entropy of the graph $G$ determined by the coding tree $T$ with a certain height $K$ is defined as:
\begin{equation}
	H_{K}(G)=\min_{\{T|height(T) \leq K\}}H^{T}(G).
\end{equation}
More details about structural entropy and coding trees are provided in Appendix~\ref{apdx:entropy}.
\paragraph{Structural Entropy Minimization.}
To minimize the structural entropy is to construct the optimal coding tree of graph $G$. Thus, we design two algorithms to heuristically construct a coding tree $T$ with height $K$. In Algorithm~\ref{alg:1}, we take $V_G$ as the leaf nodes, connect them directly to the root node $v_r^T$, and call Algorithm~\ref{alg:2} to construct an initial coding tree $T$. \footnote{Due to different scopes, we use  $T$ and $\mathbf{T}$ to distinguish the coding trees in Algorithm~\ref{alg:1} and Algorithm~\ref{alg:2}.} Algorithm~\ref{alg:2} creates a new coding tree $\mathbf{T}$ of height 1 with $\mathbf{v}$ and iteratively compresses two child nodes from the children set $C(\mathbf{v})$ of root node $\mathbf{v}$ in the first $while$ loop (lines 3-6), prioritizing the nodes that result in the largest reduction in structural entropy. Since tree $\mathbf{T}$'s height may exceed $K$, in the second $while$ loop (lines 7-10), we iteratively remove non-leaf nodes of $\mathbf{T}$, prioritizing nodes with the smallest entropy increase upon deletion. Afterward, leaf nodes of $\mathbf{T}$ might have different heights, contradicting the definition of coding trees. Thus, we adopt the operation in line 11 to align leaf nodes. Algorithm~\ref{alg:2} always returns a coding tree $\mathbf{T}$ with height 2.
Algorithm~\ref{alg:1} will iteratively invoke Algorithm~\ref{alg:2} until the height of $T$ reaches $K$. More precisely, each iteration within the $while$ loop will increment the height of $T$ by 1 by calling Algorithm~\ref{alg:2} on the root node $v_r^T$ or on all nodes in $V_T^1$, depending on the reduction in structural entropy\footnote{In the implementation, we apply pruning strategies to improve the efficiency, but they are omitted here for clarity.}. More details about the proposed algorithms can be found in Appendix ~\ref{apdx:alg}. 
\renewcommand{\thealgorithm}{\arabic{algorithm}}
\begin{algorithm}[!htb]
	\caption{Greedy Coding Tree Construction}
	\label{alg:1}
	\textbf{Input}: A graph $G=(V_G,E_G)$ and a positive integer $K$.\\
	\textbf{Output}: Coding tree $T=(V_T,E_T)$ of the graph $G$ with height $K$.\\
	\vspace{-5mm}
	\begin{algorithmic}[1] 
		\STATE $V_T^1:=\{v_r^T\}, V_T^0:=C(v_r^T) := V_G$;\\
		\STATE $T = Algo.~\ref{alg:2}(\mathbf{v}:=v_r^T)$;\\
		\WHILE{$T.height < K$}
		\STATE $T_1 = T.merge(Algo.~\ref{alg:2}(\mathbf{v}:=v_r^T))$;\\
		\STATE $T_2 = T.merge(\{ \mathbf{T} = Algo.~\ref{alg:2}(\mathbf{v}:=\hat{v})| \forall \hat{v} \in V_T^1\})$;\\
		\STATE $T =(H^{T_1}(G) < H^{T_2}(G))  \ ? \ T_1 : T_2 $;\\
		\ENDWHILE \\					
		\STATE \textbf{return} $T$;\\
	\end{algorithmic}
\end{algorithm}
\begin{algorithm}[!htb]
	\caption{2-level sub-coding tree construction.}
	\label{alg:2}
	\textbf{Input}: A node $\mathbf{v}$. \\
	\textbf{Output}: Coding tree $\mathbf{T} = (V_{\mathbf{T}}, E_{\mathbf{T}})$ with height 2.\\
	\begin{algorithmic}[1]   
			\vspace{-5mm}
			\STATE $v_r^{\mathbf{T}} := \mathbf{v}, V_{\mathbf{T}}^0 := C(\mathbf{v})$;\\
			\STATE $\forall v \in V_\mathbf{T}^0, v.parent := v_r^{\mathbf{T}}, C(v_r^{\mathbf{T}}) := {v} \cup C(v_r^{\mathbf{T}})$\\
			\label{stg:1}
			\WHILE{$|C(v_r^{\mathbf{T}})| > 2$}
			\label{stg:2}
			\STATE $(v_\alpha, v_\beta) = \underset{(v, v')}{argmax}\{H^{\mathbf{T}}(G) - H^{{\mathbf{T}}.compress(v, v')}(G)\}$
			\label{stg:3}
			\STATE $\mathbf{T}.compress(v_\alpha, v_\beta)$
			\ENDWHILE \\
			\WHILE{$\mathbf{T}.height > 2$}
			\STATE $v_i = \underset{v}{argmin}\{H^{\mathbf{T}.remove(v)}(G) - H^{\mathbf{T}}(G)\}$
			\STATE $\mathbf{T}.remove(v_i)$
			\ENDWHILE
			\STATE $\mathbf{T}.align()$
			\STATE \textbf{return} $\mathbf{T}$
		\end{algorithmic}
\end{algorithm}
\vspace{-3mm}
\paragraph{Hierarchical Representation Learning.}
After calling Algorithm~\ref{alg:1}($G_L = (V_G, E_G), K$), we get a coding tree $T_L = (V_{T_L}, E_{T_L})$ of label hierarchy $G_L$ with $T_L.height = K$ . For representation learning, reformulate the label hierarchy and its coding tree as triplets: $G_L = (V_{G_L}, E_{G_L}, X_{G_L})$, $T_L = (V_{T_L}, E_{T_L}, X_{T_L})$ where $X_{G_L} \in \mathbb{R} ^ {\mathbb{Y} \times d_V}$ is derived from document embedding $h_D$ via two-dimensional projector, formally,  $X_{G_L} = \phi_{proj}(h_D)$. $\phi_{proj}(\cdot)$ consists of a $(\mathbb{Y} \times 1) $ and a $(d_B \times d_V)$ feed-forward network, where $d_V$ is the hidden size for vertices. Meanwhile, $X_{T_L} = \{X_{T_L}^0, X_{T_L}^1, \dots, X_{T_L}^K\}$ represents the node embeddings of $V_{T_L}^i$, $i \in [0, K]$. 


In Algorithm~\ref{alg:1},  the leaf nodes $V_{T_L}^0 $ are initialized with $V_{G_L}$,  thus $X_{T_L}^0 := X_{G_L}$. Furthermore, $\{V_{T_L}^k | k \in [1, K]\}$ is given by Algorithm~\ref{alg:1} while their node embeddings $\{X_{T_L}^k | k \in [1, K]\}$ need to be fetched. Based on the structure of coding trees, we design a hierarchical representation learning module. For $x_{v}^k \in X_{T_L}^k$ in the $k$-th layer,
\begin{equation}
	x_v^k = \phi_{FFN}^k(\sum\nolimits_{n\in C(v)} x_n^{k-1}),
	\label{equ:noderep}
\end{equation}
where $v \in V_T^k$, $x_v^k \in \mathbb{R} ^ {d_V}$ is the feature vector of node $v$, and $C(v)$ represents the child nodes of $v$ in coding tree $T_L$. $\phi_{FFN}^i(\cdot)$ denotes a feed-forward network.  The information from leaf nodes propagates layer by layer until it reaches the root node.
Finally, to capture the multi-granular information provided by nodes at different levels, we utilize Equation~\ref{equ:treerep} to integrate information from each layer of $T_L$:
\begin{equation}
	\begin{split}
		h_T = \bigsqcup \limits_{k=1}^{K}{\eta(\{x_v^k|v\in V_{T_L}^k\})},
	\end{split}
	\label{equ:treerep}
\end{equation}
where $\bigsqcup(\cdot)$ indicates the concatenation operation. $\eta(\cdot)$ could be a feature-wise summation or averaging function. $h_T \in \mathbb{R} ^ {d_T}$ is the final output of the structure encoder.
\subsection{Contrastive Learning Module} 
\paragraph{Positive Pairs and Contrastive Loss.}
 We expect the text encoder and the structure encoder to learn from each other. Thus, the document embedding $h_D$ and structural embedding $h_T$ of the same sample form the positive pair in our model. Considering $h_D$ and $h_T$ might be in different distributions, we first project them into an embedding space via independent projectors, formally,
\begin{align}
		h = W_{D}^2ReLU(W_{D}^1 \cdot h_D + b_D)  \\
		\hat{h} = W_{T}^2ReLU(W_{T}^1 \cdot h_T + b_T),
\end{align}
where $h$ and $\hat{h}$ are the projected vectors of $h_D$ and $h_T$. $W_{D}^1$, $W_{D}^2$, $b_D$, $W_{T}^1$, $W_{T}^2$, and $b_T$ are weights of projectors.

Next, we utilize NT-Xent loss \citep{2020simclr} to achieve contrastive learning. Let $h_i$ and $\hat{h_i}$ denote the projected positive pair of the $i$-th document in a batch, the contrastive learning loss $L_{clr}$ is formulated as:
\begin{equation}
	L_{clr} = - log \frac{e^{\varepsilon(h_i, \hat{h_i})} / \tau}{\sum_{j=1, j \ne i}^{|\mathbf{B}|} e^{\varepsilon(h_j, \hat{h_j}) / \tau}},
\end{equation} 
where $|\mathbf{B}|$ denotes the batch size, $\tau$ is the temperature parameter, and $\varepsilon(h, \hat{h})$ is a distance metric implemented by cosine similarity $\frac{h \top \hat{h}}{||h|| \cdot ||\hat{h}||}$.
\label{sec:proof}
\paragraph{Information Lossless Contrastive Learning for HTC.}
Information lossless learning dose not imply retaining any input data.  It is preserving the minimal but sufficient information, i.e., the mutual information required by the downstream task. This is how we define information lossless learning for hierarchical text classification:
\begin{myDef}
In HTC, the mutual information between inputs and  targets can be written as:
\begin{equation}
	I((\mathcal{G_L} \circ \mathcal{D}); Y),
\end{equation}
where $\mathcal{G_L} \in \mathbb{G} , \mathcal{D} \in \mathbb{D}$ are random variables for the label hierarchy $G_L$ and the document $D$. $I(X_1; X_2)$ denotes the mutual information between random variables $X_1$ and $X_2$. $\circ$ indicates any input combination of $\mathcal{G_L}$ and $\mathcal{D}$
\end{myDef}
\begin{myDef}
	Given a function $\mathcal{F}\subseteq \mathbb{F_T} \times \mathbb{F_G}$, which is an arbitrary fusion of a text and a structure encoder. Define the optimal function $\mathcal{F}^*$ if and only if $\mathcal{F}^*$ satisfies:
	\begin{equation}
		\mathcal{F}^* = \underset{\mathcal{F}\subseteq \mathbb{F_T} \times \mathbb{F_G}}{argmax} I(\mathcal{F}(\mathcal{G_L} \circ \mathcal{D}); (\mathcal{G_L} \circ \mathcal{D})).
	\end{equation}
\end{myDef}
 That is, $\mathcal{F}^*$retains the most mutual information between the input random variables and the encoded random variables. Apparently, $\mathcal{F}^*$ is a deterministic mapping as the embedding $\mathcal{F}^*(G_L \circ D)$  is fixed for downstream prediction when given $(G_L \circ D)$. Thus, for any random variable $\xi$,
\begin{align}
	I(\mathcal{F}^*(\mathcal{G_L} \circ \mathcal{D}); \xi) = I((\mathcal{G_L} \circ \mathcal{D}); \xi).
\end{align}
When $\xi = Y$, we could have,
\begin{align}
	\label{eq:sf}
		I(\mathcal{F}^*(\mathcal{G_L} \circ \mathcal{D}); Y) = I((\mathcal{G_L} \circ \mathcal{D}); Y).
\end{align}
\begin{myThrm}
	\label{thrm:1}
	Given a document $D$ and the coding tree $T_L$ of the label hierarchy $G_L$. Denote their random variable as $\mathcal{D}$, $\mathcal{T_L}$, and $\mathcal{G_L}$. For any augmentation function  $\theta$, we have,
	\begin{equation}
		I((\mathcal{T_L} \circ \mathcal{D}); Y) \geq I(\theta(\mathcal{G_L}, \mathcal{D}); Y).
	\end{equation}
\end{myThrm}
The proof for Theorem~\ref{thrm:1} is given in Appendix~\ref{apdx:proof}.
\paragraph{Supervised Contrastive Learning.}
\label{sec:classification}
Following the training strategy of previous contrastive learning methods for HTC, we train HILL with classification loss and contrastive loss simultaneously. After executing the structure encoder, the document embedding $h_D$ and its positive view $h_T$ are concatenated and fed into the classifier along with the sigmoid function:
\begin{equation}
	P = Sigmoid([h_D; h_T] \cdot W_c + b_c),
\end{equation}
where $W_c \in \mathbb{R} ^ {(d_B + d_T) \times |\mathbb{Y}|}$ and $b_c \in \mathbb{R} ^ {|\mathbb{Y}|}$ are weights and bias of the classifier while $|\mathbb{Y}|$ is the volume of the label set. For multi-label classification, we adopt the Binary Cross-Entropy Loss as the classification loss:
\begin{equation}
	\small{L_{C} = - \frac{1}{|\mathbb{Y}|} \sum_{j}^{|\mathbb{Y}|} y_{j}log(p_{j}) + (1 - y_{j})log(1 - p_{j}),}
\end{equation}
where $y_j$ is the ground truth of the $j$-th label while $p_j$ is the $j$-th element of $P$. 

Overall, the final loss function can be formulated as:
\begin{equation}
	L = L_{C} + \lambda_{clr} \cdot L_{clr}.
\end{equation}
where $\lambda_{clr}$ is the weight of $L_{clr}$.
\begin{table}[!ht]
	\centering
	\resizebox{0.48\textwidth}{!}{
		\begin{tabular}{ccccccc}
			\toprule[1pt]
			Dataset & 
			$|\mathbb{Y}|$ &
			$Avg(|\mathcal{Y})|)$ & 
			Depth & 
			\# Train & 
			\# Dev & 
			\# Test \\ \hline
			WOS & 
			141	&	2.0	&	2 &	30,070 &	7,518 &	9,397            \\
			RCV1-v2 & 
			103	&	3.24 &	4 & 20,833 & 	2,316 & 781,265          \\
			NYTimes & 
			166 &	7.6 &	8 & 23,345 &	5,834 & 7,292            \\ 
			\hline \bottomrule[1pt]
	\end{tabular}}
	\caption{Summary statistics of the three datasets.}
	\label{tab:data_stat}
	\vspace{-2mm}
\end{table}
\begin{table*}[!th]
	\centering
	\resizebox{\textwidth}{!}{
			\begin{tabular}{lcccccccc}
			\toprule[1.5pt]
			\multicolumn{1}{c}{\multirow{2}{*}{Model}} 
			& \multicolumn{2}{c}{WOS}               
			& \multicolumn{2}{c}{RCV1-v2}           
			& \multicolumn{2}{c}{NYTimes}           
			& \multicolumn{2}{c}{Average}           \\ \cline{2-9} 
			\multicolumn{1}{c}{}                       
			& Micro-F1             & Macro-F1       
			& Micro-F1       		& Macro-F1             
			& Micro-F1             & Macro-F1       
			& Micro-F1             & Macro-F1       \\ 
			\midrule[1pt]
			\multicolumn{9}{c}{\textbf{Supervised Learning Models}}                                                                                                                                                        \\ 
			\midrule[1pt]
			TextRCNN  \citep{HiAGM}                              
			& 83.55                & 76.99          
			& 81.57          & 59.25                
			& 70.83                & 56.18          
			& 78.65                & 64.14          \\
			HiAGM \citep{HiAGM}                                     
			& 85.82                & 80.28          
			& 83.96          & 63.35                
			& 74.97                & 60.83          
			& 81.58                & 68.15          \\
			HTCInfoMax \citep{HTCInfoMax}                               
			& 85.58                & 80.05          
			& 83.51          & 62.71                
			& -                    & -              
			& -                    & -              \\
			HiMatch \citep{HiMatch}                                   
			& 86.20                & 80.53          
			& 84.73          & 64.11                
			& -                    & -              
			& -                    & -              \\ 
			\hline
			\multicolumn{9}{c}{\textbf{Supervised Learning Models (BERT-based)}}                                                                                                                                            \\ 
			\hline
			BERT $\dagger$                          
			& 85.63                & 79.07          
			& 85.65          & 67.02                
			& 78.24                & 65.62          
			& 83.17                & 70.57          \\
			BERT \citep{HiMatch}                              
			& 86.26                & 80.58          
			& 86.26          & 67.35                
			& -                    & -              
			& -                    & -              \\
			BERT+HiAGM  $\dagger$                               
			& 86.04                & 80.19          
			& 85.58          & 67.93                
			& 78.64                & 66.76          
			& 83.42                & 71.67          \\
			BERT+HTCInfoMax  $\dagger$                          
			& 86.30                & 79.97          
			& 85.53          & 67.09               
			& 78.75                & 67.31         
			& 83.53                & 71.46          \\
			BERT+HiMatch  $\dagger$                             
			& 86.70                & 81.06          
			& 86.33          & 68.66                
			& -                    & -              
			& -                    & -              \\ 
			\hline
			\multicolumn{9}{c}{\textbf{Contrastive Learning Models}}                                                                                                                                                   \\ 
			\hline
			HGCLR \citep{HGCLR}                                     
			& 87.11                & 81.20          
			& 86.49          & 68.31                
			& 78.86                & 67.96          
			& 84.15                & 72.49          \\
			\textbf{HILL(Ours)}                                       
			& \textbf{87.28} & \textbf{81.77} 
			& \textbf{87.31} &  \textbf{70.12} 
			&  \textbf{80.47} & \textbf{69.96} 
			&  \textbf{85.02} & \textbf{73.95} \\ 			
			\bottomrule[1pt]
		\end{tabular}
	}
	\caption{Experimental results of our proposed model on three datasets and their average performance. The supervised learning models (in the upper part) originally take TextRCNN \citep{Lai2015TextRCNN} as the text encoder. For fairness, we compared with their BERT-based versions implemented by \citet{HGCLR} (in the middle part).  The best results are marked in \textbf{bold}. ``-'' means not reported or not applicable. }
	\label{res:main}
	\vspace{-4mm}
\end{table*}
\section{Experiment}
\subsection{Experiment Setup}
\paragraph{Datasets and Evaluation Metrics.}
Experiments are conducted on three popular datasets in HTC. RCV1-v2 \citep{RCV1} and NYTimes \citep{NYT} consists of news articles, while WOS \citep{HDLTex} includes abstracts of academic papers. Each of these datasets is annotated with ground-truth labels existing in a predefined hierarchy. We split and preprocess these datasets following \citep{HGCLR}. The statistics of these datasets are shown in Table~\ref{tab:data_stat}. The experimental results are measured with Micro-F1 and Macro-F1 \citep{Gopal2013RecursiveRF}. Micro-F1 is the harmonic mean of the overall precision and recall of all the test instances, while Macro-F1 is the average F1-score of each category. Thus, Micro-F1 reflects the performance on more frequent labels, while Macro-F1 treats labels equally.
\paragraph{Implementation Details.}
\label{sec:detail}
For the text encoder, we use the BertModel of \texttt{bert-base-uncased} and update its parameters with Adam \citep{Adam} as the initial learning rate is 3e-5. For all three datasets, the hidden sizes $d_B ,  d_V, d_T$ are all set to 768. $\phi_{FFN}^k$ in Equation~\ref{equ:noderep} is implemented by $K$ independent multi-layer perceptions which consists of 2-layer linear transformations and non-linear functions. The batch sizes are set to 24 for WOS and NYTimes while 16 for RCV1-v2. The $\eta(\cdot) $ in Equation~\ref{equ:treerep} is implemented by a summation for WOS and NYTimes while averaging for RCV1-v2. The learning rate of the structure encoder is set to 1e-3 for WOS, 1e-4 for RCV1 and NYTimes. The weight of contrastive loss $\lambda_{clr}$ is respectively set to 0.001, 0.1, 0.3 for WOS, RCV1, and NYTimes. The optimal height $K$ of coding trees goes to 3, 2, and 3.

\paragraph{Baselines.}
We compare HILL with self-supervised only models including HiAGM\citep{HiAGM}, HTCInfoMax \citep{HTCInfoMax}, HiMatch \citep{HiMatch} and their BERT-based version and a contrastive learning model HGCLR \citep{HGCLR}. HiAGM, HTCInfoMax, and HiMatch use different fusion strategies to model text-hierarchy correlations. Specifically, HiAGM proposes a multi-label attention (HiAGM-LA) and a text feature propagation technique (HiAGM-TP) to get hierarchy-aware representations. HTCInfoMax enhances HiAGM-LA with information maximization to model the interaction between text and hierarchy. HiMatch treats HTC as a matching problem by mapping text and labels into a joint embedding space. HGCLR makes BERT learn from the structure encoder with a controlled document augmentation. 

\subsection{Results and Analysis}
The main results are presented in Table~\ref{res:main}. In the supervised-only models, HTCInfoMax enhances HiAGM through the maximization of mutual information principles. HiMatch treats HTC as a matching problem, and the reported results stand out as the best among these models. The replacement of TextRCNN with BERT has minimal impact on the relative ranking of their outcomes. This implies that the text encoder primarily influences the overall effectiveness of their models, while their specific merits are determined by their efforts beyond the text encoder. To some extent, it also indicates that their text encoder and structure encoder operate independently.

HGCLR is the inaugural contrastive learning approach for HTC. Despite the involvement of the structure encoder, the positive sample construction in HGCLR still relies on data augmentation. Conversely, our model effectively utilizes the syntactic information extracted by the structure encoder, enabling cooperation between the text encoder and the structure encoder. Specifically, the proposed HILL surpasses all supervised learning models and the contrastive learning model across all three datasets. Our model demonstrates average improvements of 1.85\% and 3.38\% on Micro-F1 and Macro-F1 compared to vanilla BERT. In comparison with HGCLR, our model achieves nearly a 2\% performance boost on both RCV1-v2 and NYTimes. Additionally, the improvement in WOS is notable, though slightly less than that observed in the other two datasets. A likely reason is that BERT is pretrained on news corpus, and the domain knowledge acquired may differ from that of paper abstracts. However, this could also indicate effective collaboration between the text encoder and the structural encoder in our framework. As the text encoder learns robust features, the structural encoder becomes increasingly powerful, and vice versa.
\begin{table}[!th]
	\resizebox{0.48\textwidth}{!}{
		\begin{tabular}{lcc}
			\toprule[1pt]
			\multicolumn{1}{c}{\multirow{2}{*}{Ablation Models}} & \multicolumn{2}{c}{RCV1-v2}                                     \\
			\cline{2-3}
			\multicolumn{1}{c}{}                  & \multicolumn{1}{l}{Micro-F1} & \multicolumn{1}{l}{Macro-F1} \\ \hline
			HILL                                  
			& 87.31                        & 70.12                        \\
			r.p. GIN \citep{GIN}                              
			& 86.48                        & 69.30                       \\
			r.p. GAT \citep{GAT}                             
			& 86.51                      & 68.12                       \\
			r.p. GCN \citep{GCN}                              
			& 86.24                       & 68.71                      \\
			r.m. $L_{clr}$                        
			& 86.51    & 68.60                      \\ 
			r.m. Algorithm~\ref{alg:1}
			& 86.67    & 67.92						\\
			\bottomrule[1pt]
		\end{tabular}
	}
	\caption{Performance when replacing or removing some components of HILL on the test set of RCV1-v2. r.p. stands for the replacement and r.m. stands for remove. The results of r.m. $L_{clr}$ and r.m. Algorithm~\ref{alg:1} are both obtained from 5 runs under different random seeds, each of which is distinguished from HILL's at a significant level of 95\% under a one-sample t-test.}
	\label{res:ablation}
	\vspace{-2mm}
\end{table}
\begin{figure*}[!htp]
	\centering
	\subfigure[WOS]{\includegraphics[width=0.3\textwidth]{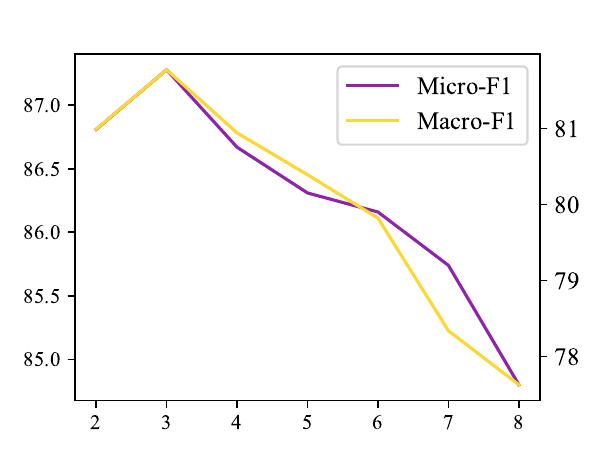}}
	\subfigure[RCV1-v2]{\includegraphics[width=0.3\textwidth]{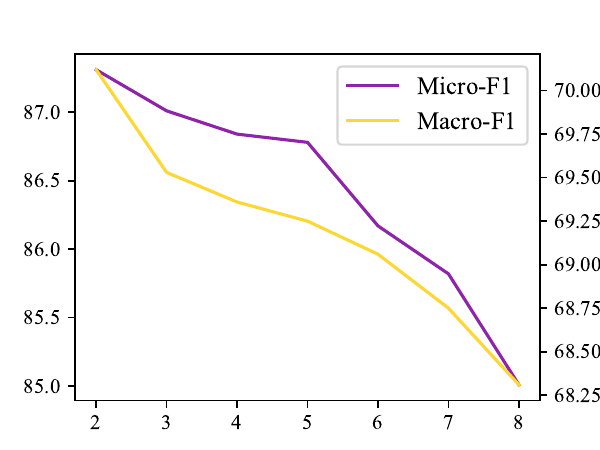}}
	\subfigure[NYTimes]{\includegraphics[width=0.3\textwidth]{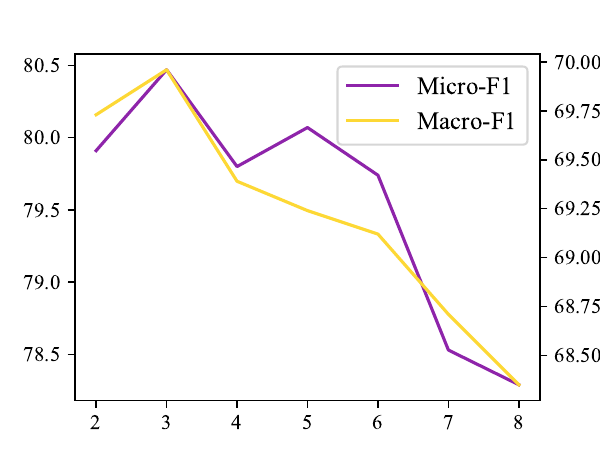}}
	\caption{Test performance of HILL with different height $K$ of the coding tree on three datasets.}
	\label{fig:k}
	\vspace{-2mm}
\end{figure*}
\subsection{Ablation Studies}
\paragraph{The necessity of proposed methods.}
\label{sec:ablation}
We conduct ablation studies by removing one component of our model at a time while keeping other conditions consistent whenever possible. The results of the ablation studies are presented in Table~\ref{res:ablation}.  To demonstrate the effectiveness of proposed hierarchical representation learning, we replaced the structure encoder with three commonly used graph neural networks including GCN \citep{GCN}, GAT \citep{GAT}, and GIN \citep{GIN}. All of them are fed with the label hierarchy $G_L$ and the document embedding $h_D$ to initialize node embeddings.  For GCN and GAT, the number of layers is set to 2 while other parameters are the default settings in PyTorch-Geometric \citep{PyG}. Regarding GIN, the combine function is a 2-layer multi-layer perception with $\epsilon=0$, and the iteration is set to 3. We find that the hierarchical learning module outperforms all graph encoders on RCV1-v2, which empirically proves that syntactic information extraction is successful in HILL. Moreover, the performance of the supervised-only model (r.m. $L_{clr}$)  declines by 0.92\% and 2.17\%, underscoring the necessity of contrastive learning. Additionally, we directly feed the initial coding tree, i.e., a coding tree within the root node $v_r$ connecting to the leaf nodes, into the structure encoder. The model (r.m. Algorithm~\ref{alg:1}) exhibits performance decreases of 0.73\% and 3.14\%, emphasizing the effectiveness of structural entropy minimization. Results and analysis on the other two datasets can be found in Appendix~\ref{apdx:ablation}.
\paragraph{The Height $K$ of Coding Trees.}
The height of the coding tree affects the performance of HILL. Higher coding trees may involve more explosive gradients. To investigate the impact of $K$, we run HILL with different heights $K$ of the coding tree while keeping other settings the same. Figure~\ref{fig:k} shows the test performance of different height coding trees on WOS, RCV1-v2, and NYTimes. As $K$ grows, the performance of HILL sharply degrades. The optimal $K$ seems to be unrelated to the heights of label hierarchies, since the heights of the three datasets are 2, 4, and 8, while the optimal $K$ is 3, 2, and 3. On the contrary, the optimal $K$ is more likely to positively correlate with the volumes of label set $\mathbb{Y}$, 141, 103, and 166.
\begin{figure}[!th]
	\centering
	\subfigure[Trainable parameters]{\includegraphics[width=0.23\textwidth]{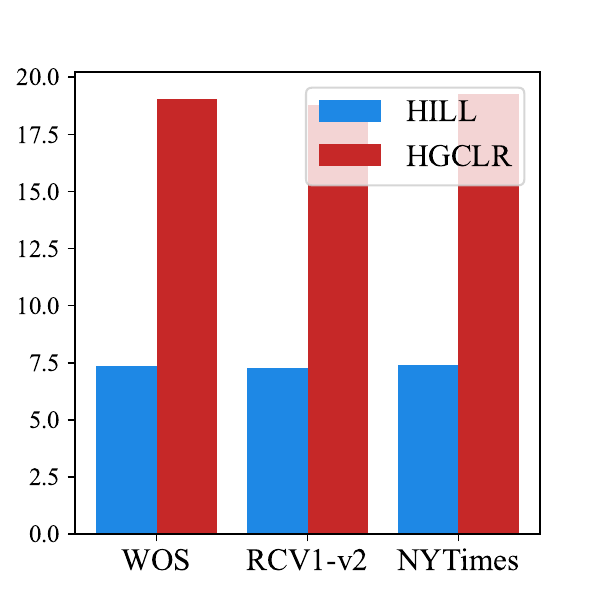}}
	\subfigure[Training time per epoch]{\includegraphics[width=0.23\textwidth]{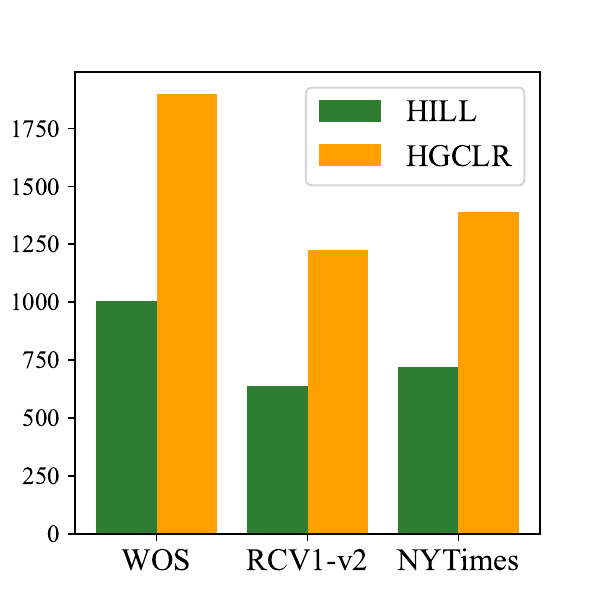}}
	\caption{The number of trainable parameters (M) and the average training time (s) of our model and HGCLR on WOS, RCV1-v2, and NYTimes.}
	\label{fig:param_time}
	\vspace{-4mm}
\end{figure}

\paragraph{Time-and-Memory-Saving Contrastive Learning.}
The structure encoder of HILL is considerably smaller than that of HGCLR, as the kernel of hierarchical representation learning consists of only $K$ multi-layer perceptions, while Graphormer \citep{Ying2021} is built upon multi-head attention. In this comparison, we assess the number of learnable parameters and the training speed of HILL in contrast to HGCLR. Specifically, we set the hidden state sizes of both HILL and HGCLR to 768, with batch sizes set to 16. The count of trainable parameters is determined by the $numel(\cdot)$ function in PyTorch \citep{Pytorch}, excluding those related to BERT. As indicated in Figure~\ref{fig:param_time}, our model exhibits significantly fewer parameters than HGCLR, averaging 7.34M compared to 19.04M. Additionally, the training speed is evaluated after 20 epochs of training\footnote{Both of them converge around the 20th epoch.}. The average training time for HILL is 789.2s, which is half the time taken by HGCLR (1504.7s). Overall, this analysis suggests that the efficient architecture of HILL contributes to its status as a time- and memory-saving model.

\section{Conclusion}
In this paper, we design a suite of methods to address the limitations of existing contrastive learning models for HTC. In particular, we propose HILL, in which the syntactic information is sufficiently fused with the semantic information after structural entropy minimization and hierarchical representation learning. Theoretically, we give the definition on information lossless learning for HTC and the information extracted by HILL is proved to be the upper bound of other contrastive learning based methods. Experimental results demonstrate the effectiveness and efficiency of our model against state-of-the-arts.

\section*{Limitations}
In Table~\ref{res:main}, we only provide the results of HILL and HGCLR when using BERT as the text encoder. Due to the focus on designing the structure encoder, we do not report results on a smaller model, for instance, TextRCNN, or a larger language model as the text encoder.  On the other hand, we adopt supervised contrastive learning in accordance with the settings of HGCLR. The performance of HILL under contrastive-supervised two-stage training remains to be explored.

\bibliography{HTC}
\bibliographystyle{acl_natbib}
\appendix
\begin{figure*}[!th]
	\centering
	\includegraphics[width=0.96\textwidth]{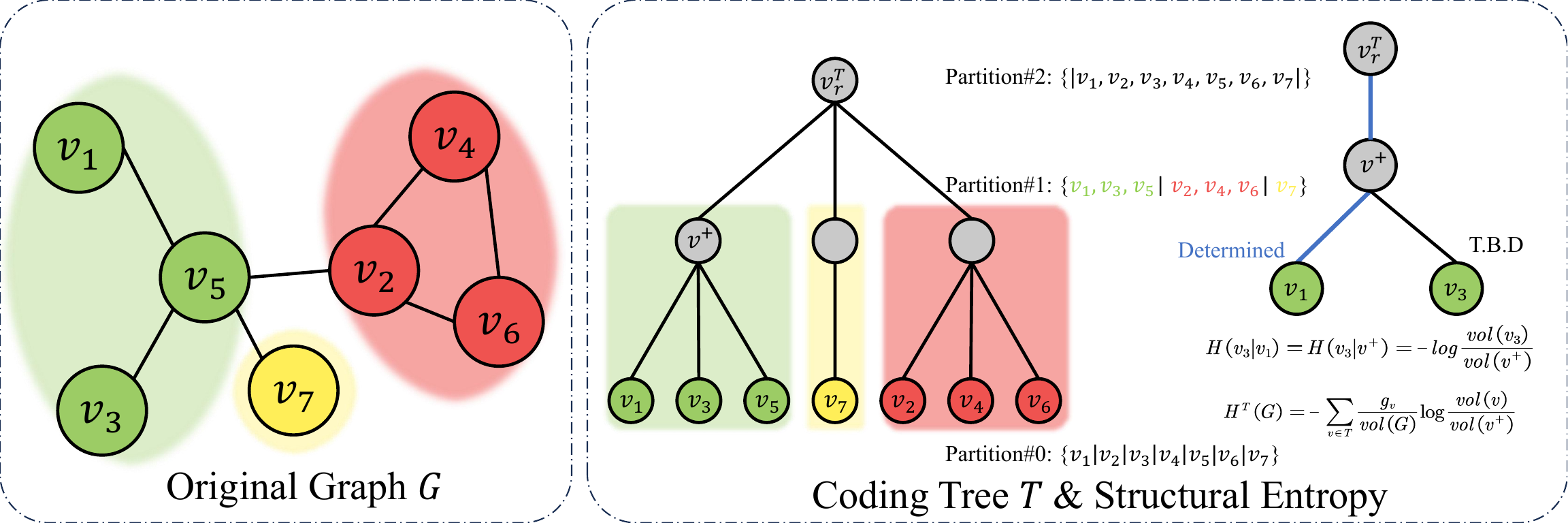}
	\caption{An illustration of coding trees and structural entropy. The coding tree $T$ provides us with multi-granularity partitions of the original graph $G$, as shown by the three partitions in the example. Structural entropy is defined as the average amount of information of a random walk between two nodes in $V_G$, considering all nodes partitioned (encoded and decoded) by coding tree $T$.  Under the guidance of structural entropy, coding tree $T$ could reveal the essential structure of graph $G$.}
	\label{fig:entropy}
	\vspace{-2mm}
\end{figure*}
\section{A Brief Introduction to Structural Entropy and Coding Trees}
\label{apdx:entropy}
The exhaustive definitions and theorems of coding trees and structural entropy are originated by \citet{Li2016StructuralEntropy}. Here, we just briefly introduce some key concepts which are crucial to this paper. An illustration of coding trees and structural entropy is shown in Figure~\ref{fig:entropy}.
\paragraph{Coding tree.} A coding tree $T = (V_T, E_T)$ of graph $G = (V_G, E_G)$ is a tree that satisfies:
\begin{enumerate}[label=\Roman*.]
	\item  Denote $v_r^T$ as the unique root node of $T$,  $V_\zeta$ as the leaf-node set of $T$, and $T.height$ the height of $T$.  The node set of $T$ is $V_T$, which consists of a few subsets, that is, $V_T = \{V_T^0, V_T^1, \dots, V_T^{T.height}\}$. $V_T^{T.height} := {v_r^T}, V_T^0 := V_\zeta$.
	
	\item Each node $v \in V_T$ is the \textbf{marker} \citep{Li2016StructuralEntropy} of a subset of $V_G$. For instance, $v_r^T$ is the marker $V_G$, while each $v_\zeta \in V_\zeta$ marks a single node in $V_G$.
	
	\item For any $V \in \{V_T^0, V_T^1, \dots, V_T^{T.height}\}$. All nodes in $V$ has the same height (depth) and any subset of $V$ represents a \textbf{partition} of $V_G$. Specifically, $V_T^0 = V_\zeta$ is an element-wise partition for $V_G$ as there exists a one-to-one correspondence for nodes in $V_\zeta$ and $V_G$. While $V_T^{T.height} = {v_r^T}$ is the overall partition for $V_G$ as $v_r^T$ marks the integrity of $V_G$. 
\end{enumerate}
\paragraph{Structural Entropy.}  As coding trees can be regarded as coding patterns for graphs, the structural entropy of a graph is defined as the average amount of information under a determined coding tree. Specifically, as depicted in Figure~\ref{fig:entropy}, when a random walk on graph $G$ progresses from node $v_1$ to node $v_3$,  a portion of the information is encoded by their parent node $v^+$, and at this point, only the information from $v+$ to $v_3$ remains to be determined. Meanwhile, the conditional entropy $H(v_3|v_1)$ is then reduced to $H(v_3|v^+)$. \footnote{Conditional entropy should be defined with random variables, but we omit them here for simplicity.} Thus, we have:
\begin{equation}
	H(v_3|v_1) = H(v_3|v^+) = - log \frac{vol\left( v_3 \right)}{vol\left( v^+ \right)},
\end{equation}
where $vol(v_3)$ denotes the degree of $v_3$ while $(v^+)$ is the total degree of its children, i.e. $vol(v^+) = vol(v_1) + vol(v_3) + vol(v_5)$. Structural entropy of a graph $G$ is defined as the average amount of information required to determine during a random walk between two accessible nodes. According  to the derivation procedure provided by \citet{Li2016StructuralEntropy}, we have:
\begin{equation}
	H^{T}(G)=-\sum_{v \in T} \frac{g_{v}}{vol(G)}\log{\frac{vol(v)}{vol(v^+)}},
\end{equation}
where $g_{v}$ represents the number of $v$'s cut edges on $G$, $vol(G)$ denotes the volume of graph $G$, and $\frac{g_v}{vol(G)}$ indicates the probability of a random walk on $G$ involving the leaf nodes marked by $v$.
\begin{figure}[]
	\centering
	\includegraphics[width=0.48\textwidth]{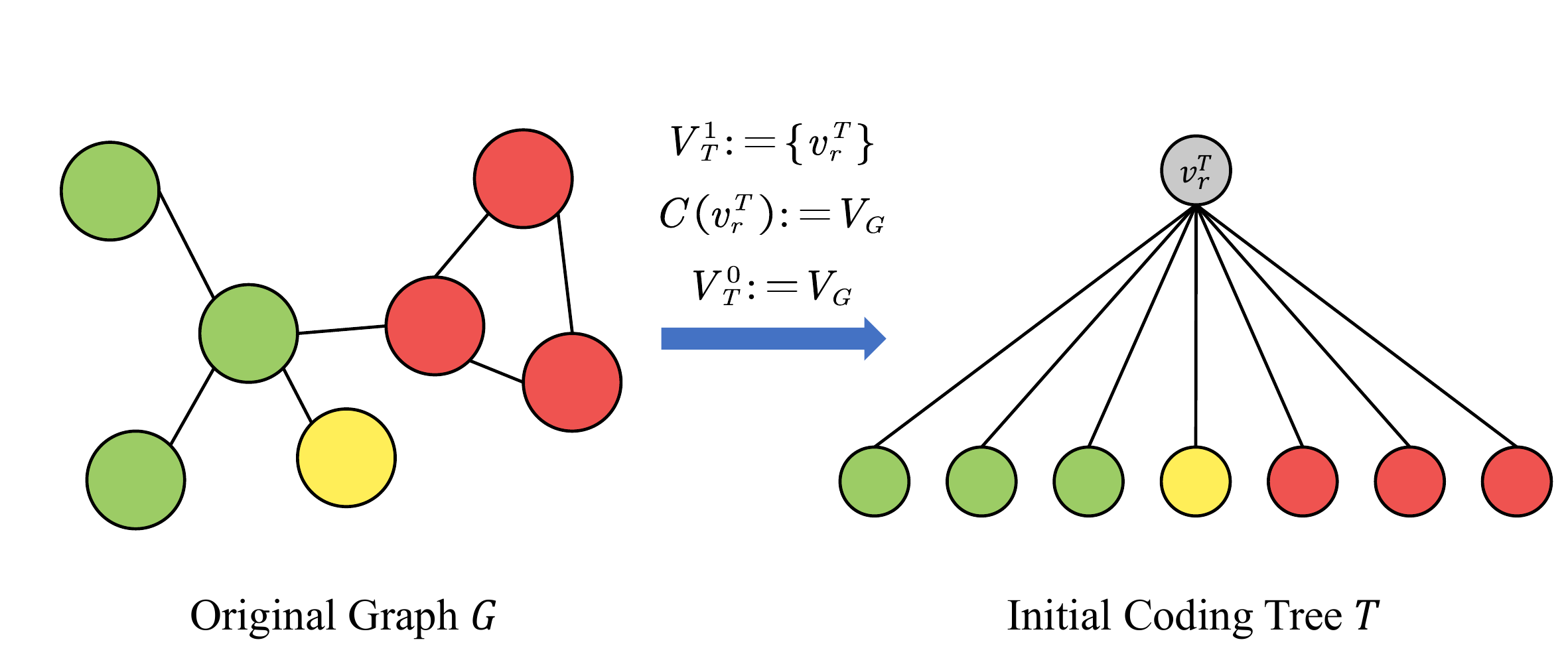}
	\caption{The initialization stage of Algorithm~\ref{alg:1}, in which a 1-height coding tree is constructed.}
	\label{fig:init}
	\vspace{-2mm}
\end{figure}
\begin{figure}[th]
	\centering
	\subfigure[An example of compressing leaf node $v_{\alpha} $ and $v_{\beta}$.\label{fig:compress1}]{\includegraphics[width=0.48\textwidth]{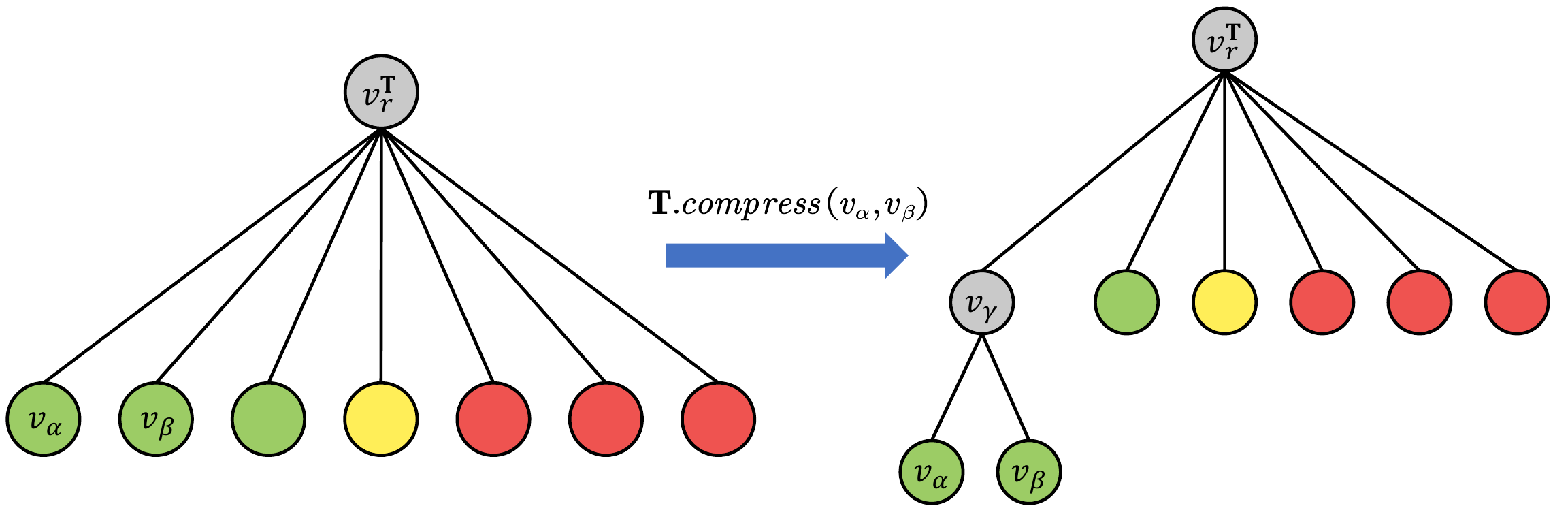}}\\
	\subfigure[The final result after executing lines 3-6.\label{fig:compress2}]{\includegraphics[width=0.48\textwidth]{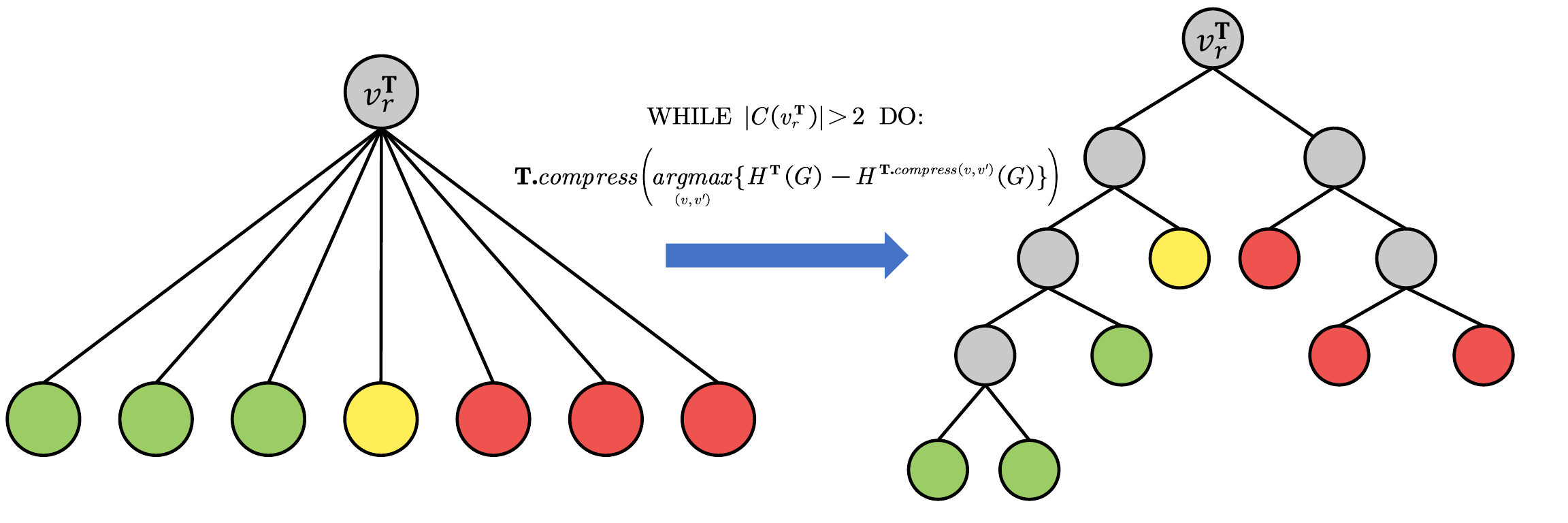}}
	\caption{An illustration of $\mathbf{T}.compress(\cdot)$ operation. (a) A single execution of $\mathbf{T}.compress(v_{\alpha}, v_{\beta})$. (b) The final state after executing lines 3-6 in Algorithm~\ref{alg:2}.}
	\vspace{-2mm}
\end{figure}
\begin{figure}[th]
	\centering
	\subfigure[An example of deleting an intermediate node $v$.\label{fig:delete1}]{\includegraphics[width=0.48\textwidth]{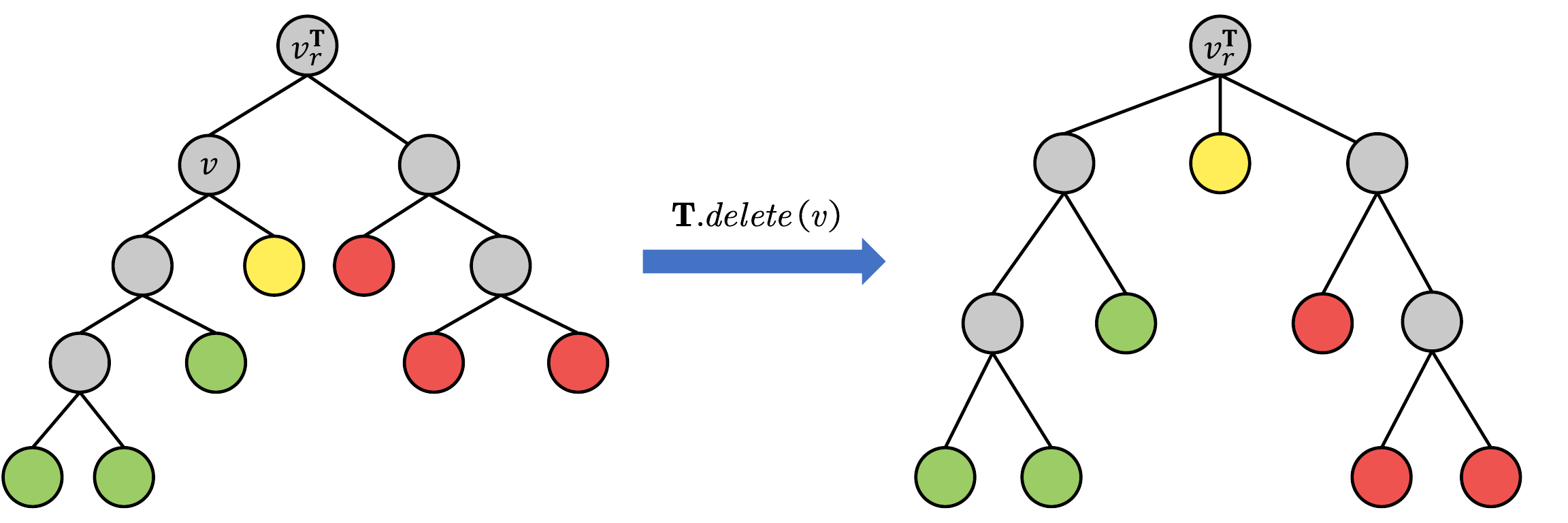}}\\
	\subfigure[The final result after executing lines 7-10.\label{fig:delete2}]{\includegraphics[width=0.48\textwidth]{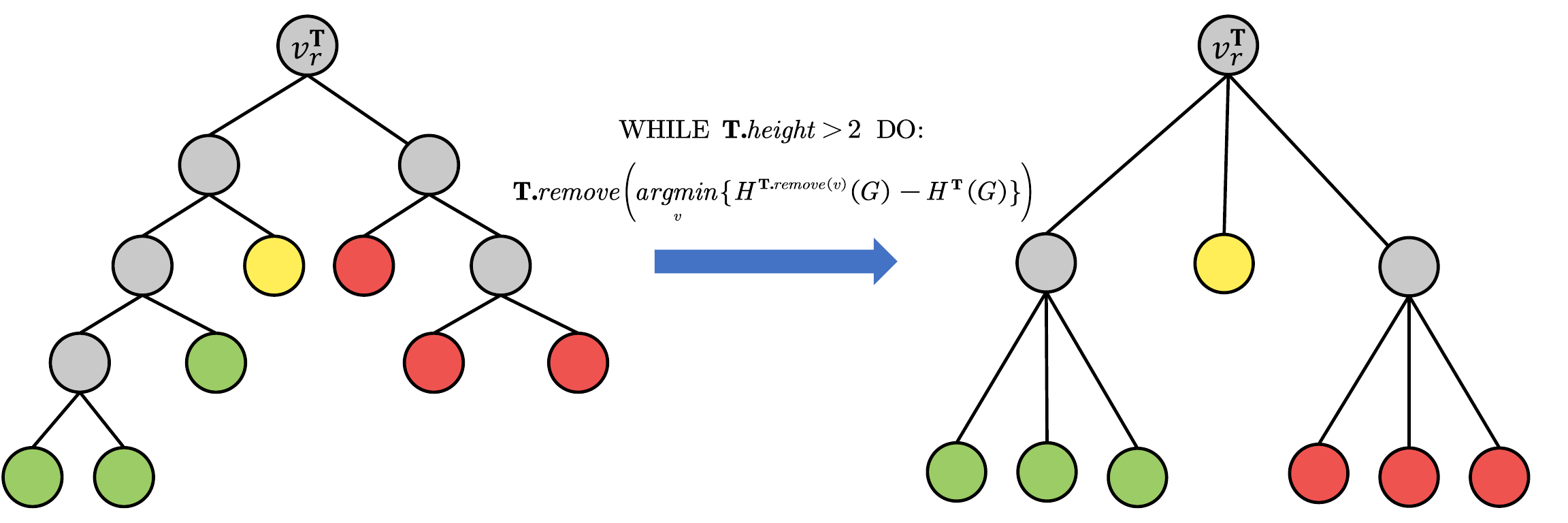}}
	\caption{An illustration of $\mathbf{T}.delete(\cdot)$ operation. (a) A single execution of $\mathbf{T}.delete(v)$. (b) The final state after executing lines 7-10 in Algorithm~\ref{alg:2}.}
	\vspace{-2mm}
\end{figure}
\begin{figure}[th]
	\centering
	\includegraphics[width=0.48\textwidth]{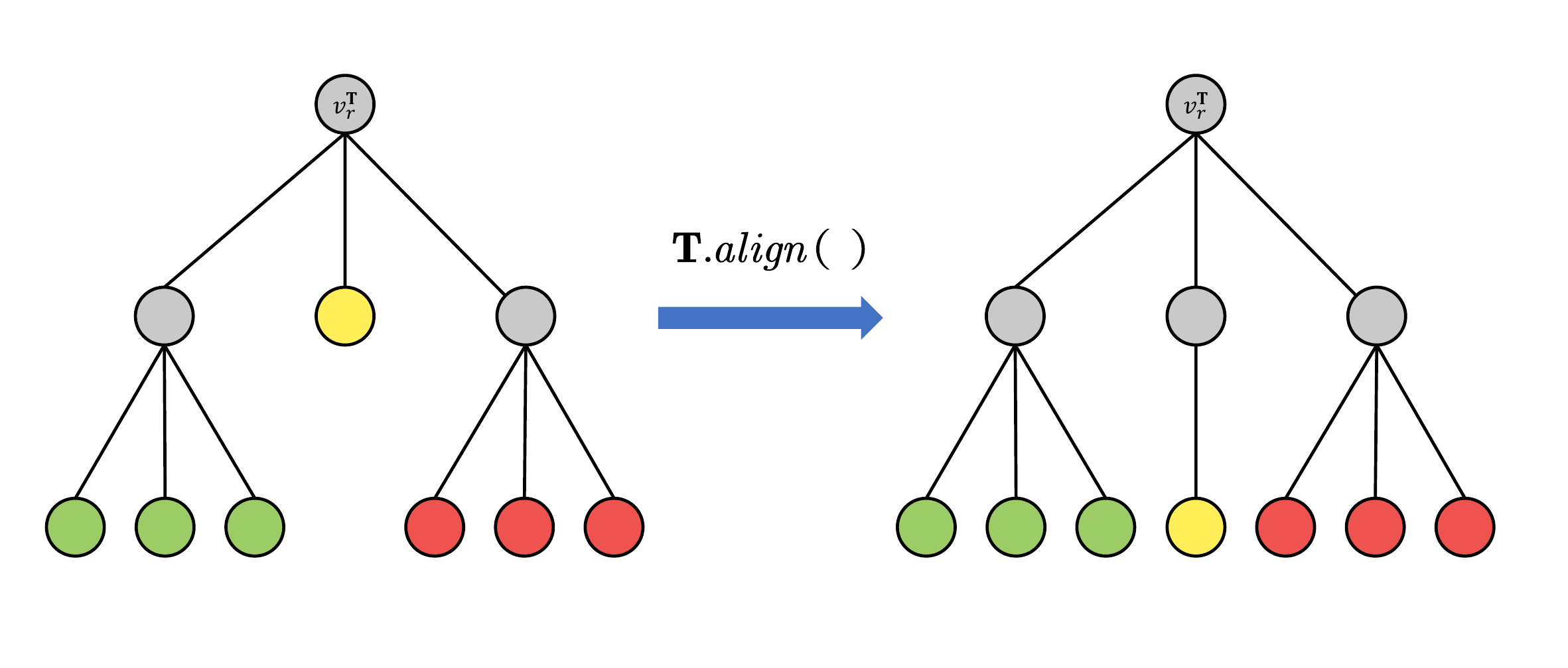}
	\caption{An illustration of $\mathbf{T}.align()$ operation. $\mathbf{T}.align()$ will align the height (depth) of all the leaf nodes to satisfy the definition of coding trees.}
	\label{fig:align}
	\vspace{-2mm}
\end{figure}
\begin{figure}[th]
	\centering
	\includegraphics[width=0.48\textwidth]{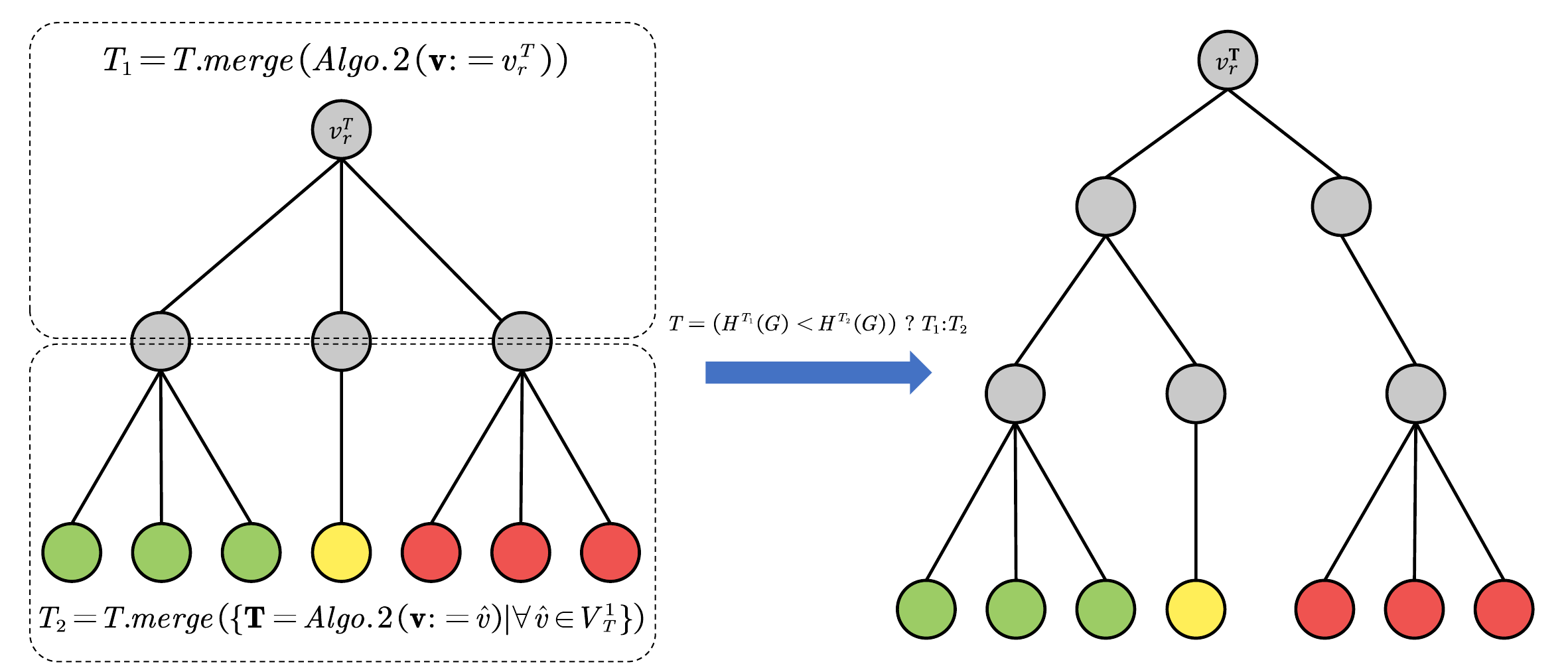}
	\caption{An illustration of lines 3-7 in Algorithm~\ref{alg:1}.}
	\label{fig:iter}
	\vspace{-2mm}
\end{figure}
\section{Explanations for The Proposed Algorithms}
\label{apdx:alg}
\paragraph{Definitions for Algorithm~\ref{alg:1} and Algorithm~\ref{alg:2}.}
Given a coding tree $T = (V_T, E_T)$, we define some attributes and member functions of $T$ as follows,
\begin{enumerate}[label=\Alph*.]
	\item Given any two nodes $v_\alpha, v_\beta$ $\in V_T$. If $(\alpha, \beta) \in E_T$,  call $v_\alpha$ the parent of $v_\beta$, and $v_\beta$ a child of $v_\alpha$, which is denoted as $v_\beta \in C(v_\alpha)$, $v_\beta.parent = v_\alpha$.
	\item Function $compress(\cdot, \cdot)$. As illustrated in Figure~\ref{fig:compress1}, given any $v_\alpha, v_\beta$ $\in C(v_r^T)$.  $compress(v_\alpha, v_\beta)$ will spawn a new node $v_\gamma$, remove $v_\alpha, v_\beta$ from $C(v_r^T)$, make $v_\gamma$ be their parent, and add $v_\gamma$ to $C(v_r^T)$. After that, $v_\alpha.parent = v_\beta.parent = v_\gamma$, and $v_\gamma \in C(v_r^T)$ while $v_\alpha, v_\beta \notin C(v_r^T)$.
	\item Function $delete(\cdot)$. As depicted in Figure~\ref{fig:delete1}, given any $v \in V_T, v \ne v_r$.  $delete(v)$ could remove $v$ from $V_T$ and attach all its children to the parent of $v$. That is, $\forall v_\mu \in C(v), v_\mu.parent := v.parent$.
	\item Function $align()$. For any leaf node $v_\zeta \in V_\zeta$, $align()$ will insert a new node between $v_\zeta$ and $v_\zeta.parent$ until the depth of $v_\zeta$ reaches $K$. $align()$ ensures that all leaf nodes of $T$ reach the same height (depth) $K$ thereby satisfying the definition of a coding tree. Figure~\ref{fig:align} shows an example of $align()$ operation.
	\item Function $merge(\cdot)$. For any node $\hat{v} \in V_T$, $Algorithm~\ref{alg:1}(\hat{v})$ returns a new coding tree $\mathbf{T}$ with height equals to 2. $T.merg(\mathbf{T})$ will replace the sub-tree of $T$ rooted by $\hat{v}$ with $\mathbf{T}$.
\end{enumerate}

Note that all the operations above update $E_T$ accordingly. In no case does $E_T$ contain self-loops or skip connections. That is, for any $(v_\alpha, v_\beta) \in E_T$, $ |v_\alpha.height - v_\beta.height| \equiv 1$.

\paragraph{Illustrations for Algorithm~\ref{alg:1} and Algorithm ~\ref{alg:2}.}
Here, we present several diagrams to deliver a running example of proposed algorithms.

As shown in Figure~\ref{fig:init}, the original graph $G$ is fed into Algorithm~\ref{alg:1} and initialized as a 1-height coding tree $T$, in which all nodes in $V_G$ are treated as leaf node and directly connect to a new root node $v_r^T$. Thereafter, Algorithm~\ref{alg:1} will call Algorithm~\ref{alg:2} several times to construct a coding tree of the specified height $K$.

Each invocation of Algorithm~\ref{alg:2} takes a non-leaf node $\mathbf{v}$ in tree $T$ as input and yields a (sub-)coding tree $\mathbf{T}$ with a height of 2 wherein $\mathbf{v}$ acting as the root. Algorithm~\ref{alg:2} first initializes $\mathbf{T}$ in a similar procedure to that illustrated in Figure~\ref{fig:init}. Subsequently, in the first $while$ loop (lines 3-6), we systematically compress the structural entropy by iteratively combining two children of root node $\mathbf{v}$ with $\mathbf{T}.compress(\cdot,\cdot)$, prioritizing those nodes resulting in the largest structural entropy reduction. Ultimately, the maximal reduction in structural entropy is achieved, resulting in the creation of a full-height binary tree, as depicted in Figure~\ref{fig:compress2}. Given that the full-height binary tree may exceed the specified height of 2, we rectify this by condensing the tree through the invocation of $\mathbf{T}.delete(\cdot)$ in the second $while$ loop (lines 7-10). The complete deletion process is illustrated in Figure~\ref{fig:delete2}. It is important to note that after condensation, tree $\mathbf{T}$ comprises leaf nodes with varying heights, which violates the definition of coding trees. To address this, in line 11, we employ $\mathbf{T}.align()$ to introduce inter-nodes. An example of $\mathbf{T}.align()$ is illustrated in Figure~\ref{fig:align}.

Once $\mathbf{T}$ returned, Algorithm~\ref{alg:1} merges $\mathbf{T}$ into $T$ at the original position of $\mathbf{v}$. Since $\mathbf{v}$ is selected from either $v_r^T$ or $V_T^1$, both of which derive sub-tree(s) with height 1, merging $\mathbf{T}$ of height 2 will increase the height of $T$ by 1. As depicted in Figure~\ref{fig:iter}, Algorithm~\ref{alg:1} aims to iteratively invoke Algorithm~\ref{alg:2} and integrate the returned $\mathbf{T}$ until $T$ reaches height $K$. The decision of whether Algorithm~\ref{alg:2} is invoked on $v_r^T$ or $V_T^1$ depends on which of the two selections results in less structural entropy for the merged coding tree $T$.
\section{Proof for Theorem~\ref{thrm:1}}
\label{apdx:proof}
\begin{myproof}
	According to \citet{Li2016StructuralEntropy},  structural entropy decodes the essential structure of the original system while measuring the structural information to support the semantic modeling of the system. Thus, we have,
	\begin{align}
		\label{eq:ess_info}
		I(\mathcal{F}^*(\mathcal{G_L} \circ \mathcal{D}); Y) =& I(\mathcal{F}^*(\mathcal{T_L} \circ \mathcal{D}); Y).
	\end{align}
	Considering the data processing inequality \citep{Cover2006} for data augmentation, we would have,
	\begin{align}
		\label{eq:dpi}
		I((\mathcal{G_L}, \mathcal{D}); Y) \geq I(\theta(\mathcal{G_L}, \mathcal{D}); Y),
	\end{align}
	where $\theta$ is a general data augmentation function acting on $(\mathcal{G_L} \circ \mathcal{D})$. Integrating the above equations, we have,
	\begin{align}
		I((\mathcal{T_L} \circ \mathcal{D}); Y) \overset{(a)}{=}&I(\mathcal{F}^{*}(\mathcal{T_L} \circ \mathcal{D}); Y) \\
		\overset{(b)}{=}& I(\mathcal{F}^*(\mathcal{G_L} \circ \mathcal{D}); Y)\\ 
		\overset{(a)}{=}& I((\mathcal{G_L} \circ \mathcal{D}); Y) \\
		\overset{(c)}{\geq} & I(\theta(\mathcal{G_L} \circ \mathcal{D})); Y).
	\end{align}
	where $(a)$, $(b)$, and $(c)$ could be referred to Equation~\ref{eq:sf}, Equation~\ref{eq:ess_info}, and Equation~\ref{eq:dpi}, respectively. Here, we have concluded the proof that the information encoded by HILL is lossless, which is the upper bound of any other augmentation-based methods. 
\end{myproof}
\begin{table*}[!th]
	\centering
	\resizebox{\textwidth}{!}{
		\begin{tabular}{lcccccccc}
			\toprule[1.5pt]
			\multicolumn{1}{c}{\multirow{2}{*}{Ablation Models}} 
			& \multicolumn{2}{c}{WOS}               
			& \multicolumn{2}{c}{RCV1-v2}           
			& \multicolumn{2}{c}{NYTimes}
			\\ \cline{2-7} 
			\multicolumn{1}{c}{}                       
			& Micro-F1 ($\Delta$)             & Macro-F1 ($\Delta$)      
			& Micro-F1 ($\Delta$)       	 & Macro-F1 ($\Delta$)             
			& Micro-F1 ($\Delta$)            & Macro-F1 ($\Delta$)
\\                                                                                                                                                     
			\midrule[1pt]
			HILL                              
			& 87.28                & 81.77          
			& 87.31          	   & 70.12                
			& 80.47                & 69.96                    
\\
			r.p. GIN \citep{GIN}                              
			& 86.67 (-0.61)               & 80.53 (-1.24)         
			& 86.48 (-0.83)               & 69.30 (-0.82)               
			& 79.43 (-1.04)               & 68.53 (-1.43)                   
\\
			r.p. GAT \citep{GAT}                                     
			& 86.58 (-0.70)                & 80.41 (-1.36)         
			& 86.51 (-0.80)               & 68.12  (-2.00)              
			& 79.46 (-1.01)               & 69.11 (-0.85)                   
\\
			r.p. GCN \citep{GCN}                               
			& 86.40 (-0.88)               & 80.47 (-1.30)         
			& 86.24 (-1.07)               & 68.71 (-1.41)               
			& 79.42 (-1.05)               & 69.33 (-0.63)             
\\
			r.m. $L_{clr}$                                   
			& 86.96 (-0.32)               & 81.37 (-0.40)         
			& 86.51 (-0.80)               & 68.60 (-1.52)               
			& 79.64 (-0.83)               & 68.90 (-1.06)             
\\ 
			r.m. Algorithm ~\ref{alg:1}                          
			& 86.21 (-1.07)               & 79.88 (-1.89)         
			& 86.67 (-0.64)               & 67.92 (-2.20)               
			& 79.63 (-0.84)               & 68.74 (-1.22)         
\\
			\bottomrule[1pt]
		\end{tabular}
	}
	\caption{Performance when replacing or removing some components of HILL on the test set of WOS, RCV1-v2, and NYTimes. r.p. stands for the replacement and r.m. stands for remove. The results of r.m. $L_{clr}$ and r.m. Algorithm~\ref{alg:1} are both obtained from 5 runs under different random seeds, each of which is distinguished from HILL's at a significant level of 95\% under a one-sample t-test. $\Delta$ denotes the decrements.}
	\label{res:ab_full}
	\vspace{-4mm}
\end{table*}
\section{Ablation Studies on WOS and NYTimes}
\label{apdx:ablation}
Table~\ref{res:ab_full} present the results of the ablation experiments on WOS and NYTimes, respectively. The experimental setups are consistent with Section~\ref{sec:ablation}. From the results, we observe that the hierarchical representation learning module proposed in this paper outperforms all other structure encoder variants on WOS \citep{HDLTex} and NYTimes \citep{NYT}. Additionally, when HILL skips Algorithm~\ref{alg:1} and completes the representation learning directly on the initial encoding tree, a significant performance drop is observed on all three datasets.

Although the results of the r.m. $L_{clr}$ for all three datasets fall short of the original HILL results, the performance gap on WOS is less pronounced compared to RCV1-v2 and NYTimes. Notably, as mentioned in Section~\ref{sec:detail}, the contrastive learning weight $\lambda_{clr}$ employed on WOS is also the smallest among the three datasets (0.001 vs. 0.1 and 0.3). However, this does not imply that introducing contrastive learning to WOS is a failure, as contrastive learning still results in substantial improvements. A more appropriate statement is that WOS reaps lower benefits due to contrastive learning. We attribute this to two factors. Firstly, there exists a substantial distributional gap between the WOS data and the pre-training data of BERT \citep{BERT}. As depicted in Table~\ref{res:main}, the performance gains achieved by replacing TextRCNN \citep{Lai2015TextRCNN} in each baseline with BERT on WOS are comparatively lower than those on the other datasets. HTCInfoMAX \citep{HTCInfoMax} even exhibits a performance degradation. Secondly, the maximum label depth of WOS is 2, significantly lower than that of RCV1-v2 (4) and NYTimes (8). This suggests that the label hierarchy of WOS contains less essential structural information, resulting in a comparatively smaller gain in the hierarchy-aware contrastive learning process. Nevertheless, the experimental results withstand the t-test, adequately demonstrating the advantages of contrastive learning.
\end{document}